# Learning sparse messages in networks of neural cliques

Behrooz Kamary Aliabadi, *Student Member, IEEE,* Claude Berrou, *Fellow, IEEE*, Vincent Gripon, *Member, IEEE*, and Xiaoran Jiang, *Student Member, IEEE*

*Abstract*—An extension to a recently introduced binary neural network is proposed in order to allow the learning of sparse messages, in large numbers and with high memory efficiency. This new network is justified both in biological and informational terms. The learning and retrieval rules are detailed and illustrated by various simulation results.

## I. INTRODUCTION

THE brain stores information with high concern for parsimony. For obvious reasons of both restricted available resource and energy limitations, the information memorized by the biological neural network results from a strong compression of the physical stimuli stemming from the "richly detailed world" [1]. For the same reasons, learning and retrieving operations involve few cerebral regions and few neurons at each time. The way the brain recruits and organizes these small populations of neurons to perform the so-called "sparse coding" of mental information [2]-[4] has still to be discovered.

On the other hand, in healthy brains, mental information is robust and durable, therefore *must be redundant*. Without redundancy, mental information would be too frail facing the physicochemical aggressions that the brain constituents suffer continuously and for so many years.

In those terms, the situation is very similar to the well-known source coding / channel coding scheme of modern telecommunication systems: firstly, information is cleared of useless components and then "intelligent redundancy" is added to allow error correction at the receiver side [5]. This rationale has recently led to the proposal of a new neural architecture combining recurrent binary networks and error correcting codes [6]. Actually, it was demonstrated in this paper that no error-correcting code had to be artificially added. Indeed, any graph, whatever the support, biological or artificial, may contain highly redundant codewords when they are assimilated to specific graph patterns, namely cliques. Exploiting this very beneficial property, multipartite clique-based networks have been proposed in [6] in order to store messages with large diversity (the number of learnable messages) and capacity (the amount of learnable binary information), as well as strong robustness towards erasures or errors. However, these networks were devised in such a way that all the clusters resulting from multipartition are used in the memorization procedure. Therefore, they do not correspond directly to the "sparse coding" vision of mental information. Moreover, the diversity of these networks is proportional to the square of the number of neurons in each cluster, and not to the square of the total number of neurons.

In order to lift these restrictions, the clique-based networks have to be reconsidered and reassessed with respect to the storage of sparse messages, that is, messages that do not call for the complete network but only for parts of it. This is the topic of this paper, the rest of which is organized in seven sections. Section II recalls the principles and notations of the clique-based networks and also proposes a slight improvement of the retrieving algorithm introduced in [6]. Some considerations about the biological plausibility of the architecture are propounded as well. In Section III, the learning and retrieving algorithms of sparse messages are described. Sections IV and V provide some theoretical analysis and simulation results for basic applications of associative memory and classification. Section VI gives an illustration of the network performance, in terms of error correction for classification. In section VII, the question of sparse messages with variable degrees of sparsity is taken up. Finally, some comments about the openings of this new kind of neural networks are proposed in the conclusion.

## II. NETWORKS OF NEURAL CLIQUES

### A. Summary

Consider a network with $n$ binary nodes linked by binary edges (that is, each edge exists with weight 1 or does not exist). This network can then be described by a non weighted, non oriented graph whose nodes may be activated or not, with respective values 1 and 0. The network is split into $c$ clusters, each containing $l = n/c$ nodes. For reasons that will be given later, these nodes are called *fanals*. Though any alphabet with cardinality $l$ could be considered in the representation of information stored by the network, we focus on binary messages in order to allow classic computations or estimations of storage properties. Therefore, $l$ is taken to be a power of 2: $l = 2^{\kappa}$. With an input binary message $m$ of length $k = c\kappa$ to

This work was supported in part by the European Research Council (ERC-AdG2011 290901 NEUCOD).

The authors are with the Electronics department of Télécom Bretagne (Institut Mines-Télécom), CS 83818, 29238 Brest, France. They are also with the Laboratory for Science and Technologies of Information, Communication and Knowledge, CNRS Lab-STICC, Brest, France (e-mail: behrooz.kamaryaliabadi@telecom-bretagne.eu; claude.berrou@telecom-bretagne.eu; vincent.gripon@telecom-bretagne.eu; xiaoran.jiang@telecom-bretagne.eu).

learn, is associated a unique set of fanals, one per cluster, using the mapping:

$C : m = \{m_1, \ldots m_i, \ldots m_c\} \rightarrow (f(m_1) \ldots f(m_i) \ldots f(m_c))$

where
- $m_i$ of length $\kappa$ bits is the sub-message or character associated with the $i^{\text{th}}$ cluster,
- $f$ is the function that maps each sub-message to a unique fanal in the corresponding cluster.

Thus, the network learns a given message by selecting one fanal per cluster and connecting these $c$ fanals to build a fully interconnected subgraph, that is, a clique. In other words, learning the particular message $m$ is equivalent to learning the pattern $C(m)$. If $\mathcal{M}$ is the set of messages learnt by the network, $\mathcal{W}(m)$ the set of edges that have to be created to store particular message $m$, the ensemble $\mathcal{W}$ of existing edges resulting from the learning of $\mathcal{M}$ is simply given by

$$\mathcal{W} = \bigcup_{m \in \mathcal{M}} \mathcal{W}(m). \quad (1)$$

This result does not depend on the order in which messages are presented, and learning a new message can be done at any moment. This very simple learning rule leads to a completely binary network. Note that no connection is established within a cluster.

The retrieving algorithm is a two-step, possibly iterative, procedure. First, at the global scale and from what is known of the stimulus (that is, some of the sub-messages $m_i$), the corresponding fanals send unitary signals towards the network through established connections and then, the contributions are added at each node. After this message passing step, at the local scale of each cluster, a winner-take-all rule is performed. Noting $v(n_{ij})$ the value of the $j^{\text{th}}$ fanal in the cluster with index $i$ ($1 \leq i \leq c; 1 \leq j \leq l$) and $w_{(i'j')(ij)}$ the weight (0 or 1) of the edge between fanals $n_{i'j'}$ and $n_{ij}$, the global decoding equation is

$$v(n_{ij}) \leftarrow \sum_{i'=1}^{c} \max_{1 \leq j' \leq l} \left( w_{(i'j')(ij)} v(n_{i'j'}) \right) + \gamma v(n_{ij}). \quad (2)$$

A memory effect, with parameter $\gamma$, is added to the message passing procedure. This relation is slightly different from the one proposed in [6], as the summation on node $n_{ij}$ of the signals stemming from the same cluster with index $i'$ is replaced with a selection of its maximum. The reason why equation (2) is now preferred is detailed in [7]. Briefly, the max function is justified by the following argument: when several fanals are active within the same cluster, that is, in presence of ambiguity, this very cluster must not impact on the rest of the network more than if there were only one active fanal. In other words, ambiguity is tolerated at the local level but not favored at the global scale.

As for the local "winner-take-all" selection, the relations are the same as in [6], precisely:

$$\forall i, 1 \leq i \leq c : v_{\max,i} \leftarrow \max_{1 \leq j \leq l} (v(n_{ij}))$$

$$\forall i \text{ and } j, 1 \leq i \leq c, 1 \leq j \leq l: \quad (3)$$

$$v(n_{ij}) \leftarrow \begin{cases} 1 & \text{if } v(n_{ij}) = v_{\max,i} \text{ and } v_{\max,i} \geq \sigma \\ 0 & \text{otherwise} \end{cases}.$$

After these operations, all fanals have value 1 or 0, which explains why this network is said to be binary, even if transitory fanal values may be larger than 1. $\sigma$ is a threshold which is quite comparable to that of the McCulloch-Pitts model of neuron [8] and which may be used as an additional level of control. In normal conditions and classical applications, all the $v_{\max,i}$ computed in the $c$ clusters are equal and can then be reduced to a single maximal score $v_{\max}$. A counter-example would be, for instance, a network in which some established connections have disappeared, due to some physical flaw; in these conditions, some signals may be missing in the computation described by (2), preventing one or more fanal values from reaching $v_{\max}$.

After the calculations formulated by (3), more than one fanal may remain activated within the same cluster. In this ambiguous situation, a repetition of (2) and (3) may be profitable and the process may need several iterations to converge towards a non-ambiguous fixed point.

*B. Biological considerations*

From the point of view of informational organization, the fundamental processing unit in the brain is likely to be the microcolumn (also called the minicolumn) [9]-[11]. This very heterogeneous group of about 100 neurons repeats itself quasi-uniformly over the about 20 square decimeters of the human grey matter. Therefore, the microcolumn, as an "identical repeating unit", can be seen as a node in a graph, able to receive and send many signals from and towards the rest of the network, with the same informational abilities everywhere in the cortex. As said in [11]: "current data on the microcolumn indicate that the neurons within the microcolumn receive common inputs, have common outputs, are interconnected, and may well constitute a fundamental computational unit of the cerebral cortex".

These microcolumns are grouped into columns whose populations are various, from some tens to some hundreds. These columns are believed to contain microcolumns that react to the same family of stimuli (e. g. the value of an angle in the visual cortex). Rising in the neural hierarchy, columns gather in macrocolumns and then several macrocolumns together constitute the so-called functional areas of the brain.

We propose to liken the fanals of the clique-based networks to microcolumns, the clusters to columns, and finally the network itself to a macrocolumn. For the sake of simplicity, all the clusters have same cardinality $l$ in the paper but as already stated, biological columns are of various sizes. The term *fanal* has been adopted to represent a node in the clique-based networks for two reasons. First, it expresses the reality of a cluster in which only one fanal can be "lighted" during the learning process; second, it says that a node of the graph is not a single neuron, but a group of neurons, namely a microcolumn.

## III. PROCESSING SPARSE MESSAGES

Consider a message of length $\chi$, composed of characters taken in an alphabet denoted by $\mathcal{A}$, the neutral character 0 being one of its elements. According to the classical definition, a "sparse message" is a message containing few non-0 characters. Such messages are the subject of many current studies in the field of information theory, especially for the "compressed sensing" application [12], or also for classification [13]. We have here to change the definition and say that a message is sparse *when it contains few expressed characters*, that is, a limited number of significant characters, in specific locations, the others being of no concern. To take a simple example, if $\mathcal{A}$ is the ensemble of positive or null integers less than 8, a sparse message of length $\chi = 24$, according to the conventional definition, could be: (000500060010000030000000), whereas we consider here messages such as: (---5---6--1-----3-------), each dash meaning "no-care". The non-expressed characters do not need to be stored by the neural network and do not require any fanal to be recruited in the learning of the message, while a classical sparse message would need the writing of the 0s. This vision of sparse messages is much more in accordance with the way the brain learns its knowledge elements, with the concern for parsimony.

*A. Learning*

Until section VII, we consider the learning of messages of length $\chi$ with the same number $c$ of few expressed characters that we call indifferently the *clique order* or *message order* in the sequel. Alphabet $\mathcal{A}$ contains $l = 2^\kappa$ elements and, in the same way as proposed in [6], the network is split into $\chi$ clusters, each containing $l$ fanals. Therefore, the network contains $n = \chi l$ fanals and $\frac{\chi(\chi-1)l^2}{2}$ potential connections, that is, a binary resource of

$$Q = \frac{\chi(\chi-1)l^2}{2} \text{ [bits]}. \quad (4)$$

Starting from an initial state with no connection at all, the learning of a first message will recruit $c$ fanals and establish $\frac{c(c-1)}{2}$ connections to build up a clique (see an example in Fig. 1 for $\chi = 9$, $l = 16$ and $c = 4$ and one established clique with $\frac{4 \times 3}{2} = 6$ connections). The probability that any particular connection does not belong to this clique is supposed to be: $1 - \frac{c(c-1)}{\chi(\chi-1)l^2}$ (this is not rigorously exact because the connections created by the clique are correlated, but the effect of this correlation is negligible when a large number of cliques have been formed). So, after the learning of $M$ independent and identically distributed (i.i.d.) random messages, this probability, which we assume to give directly the expected connection density, is

$$d = 1 - \left(1 - \frac{c(c-1)}{\chi(\chi-1)l^2}\right)^M. \quad (5)$$

Reciprocally, given a density $d$, the number $M$ of i.i.d. learned messages is

$$M = \frac{\log(1-d)}{\log\left(1 - \frac{c(c-1)}{\chi(\chi-1)l^2}\right)}. \quad (6)$$

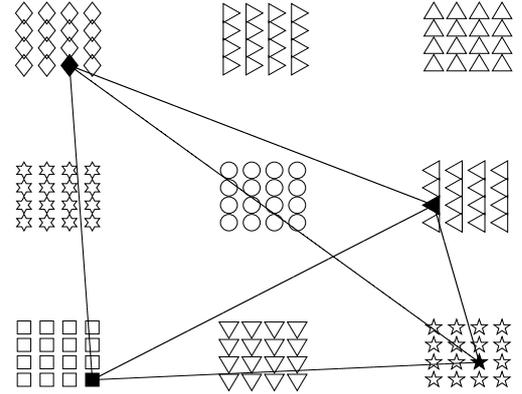

Fig. 1. Network composed of 9 clusters of 16 fanals each. A first clique with 4 vertices has been formed in the network.

Fig. 2 shows the evolution of $d$ as a function of $M$, for $\chi = 100$, $l = 64$ and various values of $c$. For low values of $d$, we have

$$d \approx \frac{c(c-1)M}{\chi(\chi-1)l^2} \quad (7)$$

and reciprocally

$$M \approx \frac{\chi(\chi-1)l^2}{c(c-1)}d \approx \frac{\chi^2 l^2}{c(c-1)}d = \frac{n^2 d}{c(c-1)} \quad (8)$$

for $\chi \gg 1$. This very simple result shows that, for a given density $d$ and a particular value of the clique order, the number of messages that the network is able to learn is proportional to the square of the number of nodes, or fanals. This quadratic law has for instance to be compared to the well-known sub-linear law of Hopfield networks (see [6] for details) or other comparable laws obtained with networks based on weighted connections.

We can express the amount of binary information $B$ learned by the network after the memorization of $M$ messages as

$$B = M\left(\log_2\left(\binom{\chi}{c}\right) + c\kappa\right). \quad (9)$$

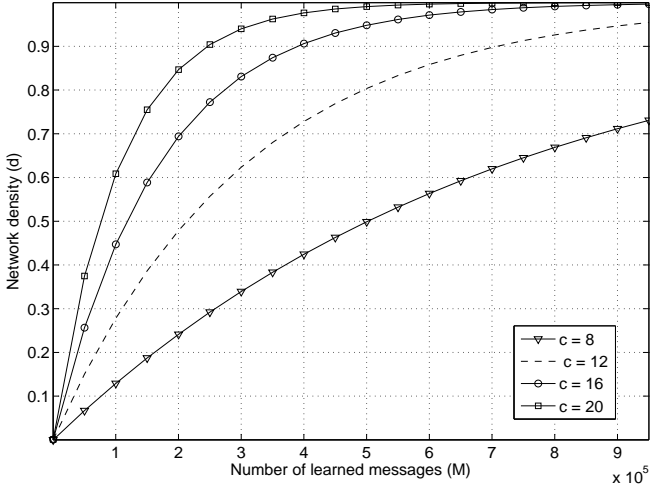

Fig. 2. Density of the network connections as a function of the number $M$ of learned messages, for $\chi = 100$, $l = 64$ and four values of $c$.

The first term between the parentheses accounts for the choice of the clusters in the storage of one message, with $c$ expressed characters among $\chi$. The second term represents the binary content of each message where $\kappa = \log_2(l)$. Finally, we define efficiency $\eta$ as the ratio of $B$ and $Q$:

$$\eta = \frac{B}{Q} = \frac{2M\left(\log_2\binom{\chi}{c} + c\kappa\right)}{\chi(\chi-1)l^2}. \quad (10)$$

For efficiency equal to 1, this formula leads to an upper bound for $M$, called the *efficiency-1 diversity* of the network[1]:

$$M_{\max} = \frac{\chi(\chi-1)l^2}{2\left(\log_2\binom{\chi}{c} + c\kappa\right)}. \quad (11)$$

For instance, with $\chi = 100$, $l = 64$ and $c = 16$, $M_{\max}$ is around $1.30 \times 10^5$ messages of 156 bits each. Density calculated from (5) is then about 0.54.

From (11), we observe that the largest values of $M_{\max}$ are obtained for the lowest clique orders $c$. This is quite natural: for a given binary resource, the shorter the messages are, the more numerous they can be. But the smaller a clique is, the less redundant and robust it is facing possible flaws (e.g. errors, erasures or over learning). The choice of $c$ is thus the result of a trade-off between diversity and robustness. The optimal values for the applications of associative memory will be discussed in section IV.

*B. Retrieving*

The basic equations for the retrieving of learned messages are still (2) and (3) but in which some adjustments have to be made to take sparsity into account:

---
[1] However, because messages are not ordered in an associative memory and therefore require less resource than ordered messages, efficiency larger than 1 is not inconceivable [6].

$$v(n_{ij}) \leftarrow \sum_{i'=1}^{\chi} \max_{1 \le j' \le l}\left(w_{(i'j')(ij)} v(n_{i'j'})\right) + \gamma v(n_{ij}) \quad (12)$$

$$\forall i, 1 \le i \le \chi : v_{\max,i} \leftarrow \max_{1 \le j \le l}(v(n_{ij}))$$

$$v_{\max} \leftarrow \max_{1 \le i \le \chi}(v_{\max,i})$$

$$\forall i \text{ and } j, 1 \le i \le \chi, 1 \le j \le l :$$

$$v(n_{ij}) \leftarrow \begin{cases} 1 & \text{if } v(n_{ij}) = v_{\max} \text{ and } v_{\max} \ge \sigma_i \\ 0 & \text{otherwise} \end{cases}. \quad (13)$$

Let us first point out that, despite the fact that a learned message uses a restricted number $c$ of clusters at each time, these clusters vary from one message to another. Consequently, in common situations, the decoding procedure has to address the whole network uniformly. Therefore, $c$ is replaced with $\chi$ as the upper index of the summation in (12). Secondly, the winner-take-all rule expressed by (3) for each cluster has now to be extended to the whole network in order to find the appropriate clusters in the search for a particular message, if their indices are not known. To achieve this, equation (13) retains the maximum value $v_{\max}$ of maximum scores obtained in all clusters and assigns $v_{\max}$ as the condition to reach for all fanals to be active. Finally and in general cases, threshold $\sigma$ is now cluster-dependant and thus is denoted $\sigma_i$. Indeed, in some applications, all the clusters may not be involved in a specific task. For instance, if the network is asked to recognize, or reject, a particular message with $c < \chi$ expressed characters, only the corresponding clusters have to be processed. The threshold of the other clusters is then set at a high unreachable value. In another situation, if the network has to store, not only binary messages, but also some kind of cognitive features, the choice of distinct values for $\sigma_i$ may orientate the network decoding towards constrained solutions. A high value for the threshold of a particular cluster would mean that this has no relevance in the current process.

## IV. ASSOCIATIVE MEMORY

Suppose that a network with parameters $\chi$ and $l$ has learnt $M$ i.i.d. random messages with order $c$, that is, cliques with $c$ vertices. We want to know what the error probability is in the recovery of one learned message when $c_e < c$ clusters are not provided with information. Two extreme cases have to be considered. Either the indices of the $c_e$ missing clusters are completely unknown - we call this most severe case *blind recovery* -, or they are totally known - it is then called *guided recovery*.

*A. Blind recovery*

The first and most obvious reason why the retrieving of a previously learned message may fail is because at least one other learned message shares the same known characters. For i.i.d. messages and large enough values of $c - c_e$ and $l$, this probability is negligible. The second reason is the possible existence of spurious cliques that would interfere in the retrieving process. The most probable and most impeding spurious clique has $c - c_e + 1$ vertices, that is only one more than the known characters, and consequently the known clusters (the appendix gives an example of why such a spurious clique makes the decoding fail, even in an iterative process). Given the density $d$ of the network connections, the probability that a particular fanal does not form a clique of minimal order with the $c - c_e$ known fanals is $1 - d^{c-c_e}$. The number of fanals available to form such a spurious clique is $c_e(l-1) + (\chi - c)l$. Therefore, the probability that no such clique exists, which we liken to the probability $P_r$ of retrieving the correct message, is

$$P_r = \left(1 - d^{c-c_e}\right)^{c_e(l-1)+(\chi-c)l}. \quad (14)$$

The probability of error $P_e$ is then given by

$$P_e = 1 - P_r = 1 - \left(1 - d^{c-c_e}\right)^{c_e(l-1)+(\chi-c)l} \quad (15)$$

which can be approximated, for small values of $d$, as

$$P_e \approx \left(c_e(l-1) + (\chi - c)l\right) d^{(c-c_e)}. \quad (16)$$

Note that, even without erasure ($c_e = 0$), the error probability in blind recovery is not zero. This is because a given learned clique of order $c$ may belong to a spurious clique with order $c + 1$, with approximated probability $(\chi - c)ld^c$.

The curve (a) of Fig. 3 shows the simulated error rate in the blind recovery of messages, as a function of their number $M$, for $\chi = 100$, $l = 64$, $c = 12$ and $c_e = 3$, after one iteration of (12) and (13). All thresholds $\sigma_i$ are equal to zero in the decoding process and memory parameter $\gamma$ is equal to 1. Values given by (15) are also indicated, showing good correspondence between theory and simulation.

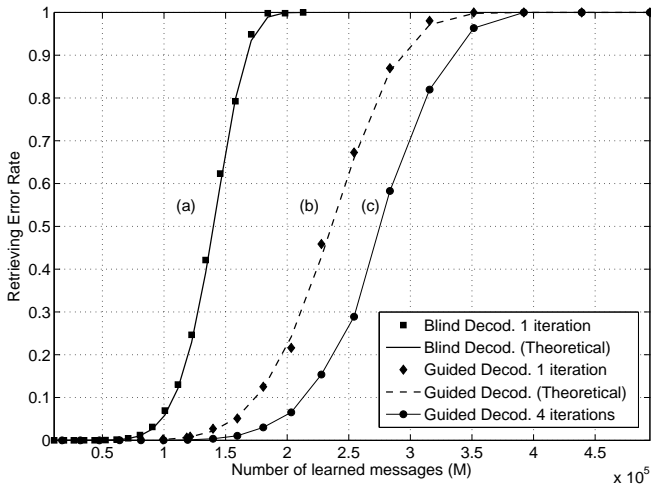

Fig. 3. Error rate in both blind and guided recovery of $M$ i.i.d. messages with order $c = 12$ in a network composed of $\chi = 100$ clusters of $l = 64$ fanals each. $c_e = 3$ clusters have no initial information.

From relations (5) and (15), it is possible to link $M$ and $c$, for a predetermined value of $P_e = P_0$. To do this easily, we use the approximations given by (7) and (16), with the additional and rough hypothesis: $\chi \gg c \gg 1$. We also set $c_e$ as a fraction $\alpha$ of $c$: $c_e = \alpha c$. All this results in

$$M \approx \left(\frac{\chi l}{c}\right)^2 \left(\frac{P_0}{\chi l}\right)^{\frac{1}{(1-\alpha)c}}. \quad (17)$$

Fig. 4 depicts the variation of diversity $M$, as a function of $c$, for $\chi = 100$, $l = 64$, $\alpha = 0.25$ and three values of $P_0$. These curves show that there exists a value of $c$ that maximizes the diversity of the network, for a given value of $P_0$. This value, denoted $c_{opt}$, is estimated by derivating (17) (through the logarithm) with respect to $c$, and finding the condition for extremum. The computation gives:

$$c_{opt} \approx \frac{\log\left(\frac{\chi l}{P_0}\right)}{2(1-\alpha)}. \quad (18)$$

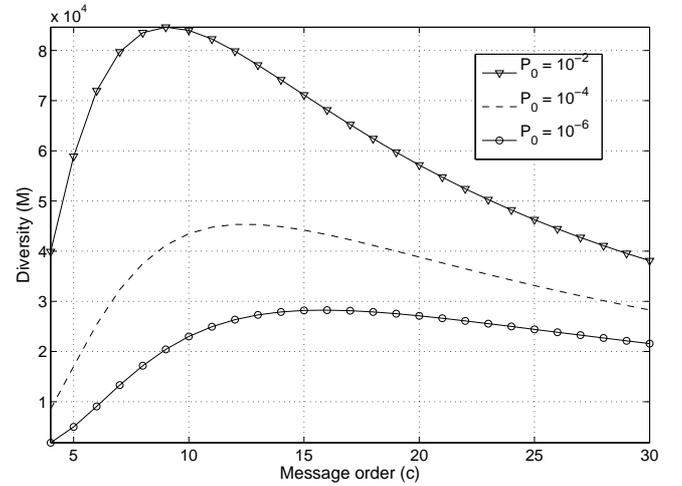

Fig. 4. Diversity of the network composed of $\chi = 100$ clusters of $l = 64$ fanals each, as a function of the message order $c$, when a quarter of the $c$ clusters have no information at the initialization step (relation (17) with $\alpha = 0.25$) and for a given error probability $P_0$ for blind recovery.

For instance, with $\chi = 100$, $l = 64$, $\alpha = 0.25$, we have $c_{opt}$ equal to 9 and 15 for $P_0 = 10^{-2}$ and $10^{-6}$ respectively. The corresponding diversities given by (17) are around 70000 and 25000 and efficiencies, as formulated by (10), are 33% and 18%. As can be surmised, material efficiency ($\eta$) and effectiveness in restoring the messages ($P_e$) are conflicting, but not so sharply.

### B. Guided recovery

In this favorable case, some characters of the message are missing but their supporting clusters are known. The thresholds $\sigma$ of irrelevant clusters are then set to an unreachable value. The situation comes down exactly to that described in [6, section VI] in which all clusters are systematically known to retrieve a message. The error probability, after one iteration, is given by

$$P_e = 1 - \left(1 - d^{c-c_e}\right)^{(l-1)c_e}. \quad (19)$$

The simulated error rate, after several iterations may be notably less than $P_e$. The curves (b) and (c) of Fig. 3 depict the evolution of the error rate as a function of $M$, after one and four iterations respectively and with the same parameters as curve (a). The theoretical values given by (19) are also indicated.

The gap in performance between blind and guided recovery is not considerable, in terms of diversity. When guided, instead of blind, decoding is performed and for a given error rate, diversity $M$ (which is roughly proportional to density $d$, through approximation (7)) is higher by about 50%, after one iteration and 100% after four iterations.

## V. CLASSIFICATION

As in [6, section V], we consider a simple application of go/no-go classification. The network with parameters $\chi$ and $l$ learns $M$ i.i.d. messages with same order $c < \chi$. The decoding procedure is still described by relations (12) and (13) with $\gamma = 1$, but thresholds $\sigma$ are now equal to $c$ for the clusters under test (those which have a fanal activated) and a higher unreachable value for the others, in order to prevent them from interfering. The network is then asked whether a message is known by it or not, that is practically, whether the decoding procedure will accept the stimulus without modification or not. The recognition of learned messages is always successful, since all the activated fanals will obtain the maximal score (i.e. $c$) as soon as the first iteration has finished and no other one can be the winner within the clusters under test. Therefore, the type I error is naught:

$$P_{\text{type I error}} = 0. \quad (20)$$

Because the thresholds of the irrelevant clusters are set to unreachable values, the only possibility for the network to accept a wrong message, that is, a type II error, is the existence of a spurious clique of order $c$ in the clusters under test. Assuming again that not only messages, but also connections are i.i.d., the probability of having such a spurious clique is

$$P_{\text{type II error}} \approx d^{\frac{c(c-1)}{2}} = \left(1 - \left(1 - \frac{c(c-1)}{\chi(\chi-1)l^2}\right)^M\right)^{\frac{c(c-1)}{2}}. \quad (21)$$

Fig. 5 represents the type II error rate obtained from simulation, on the one hand, and relation (21), on the other hand, as a function of density $d$. The parameters are $\chi = 100$, $l = 64$ and $c = 6$ or 9. The figure also indicates the values of efficiency-1 diversity $M_{\max}$, as calculated from (11), showing that low error rates may be obtained even for efficiencies higher than 1. For instance, with $c = 9$, diversities more than three times $M_{\max}$ ($d = 0.68$ according to (5)) can be attained while keeping the error rate below $10^{-6}$.

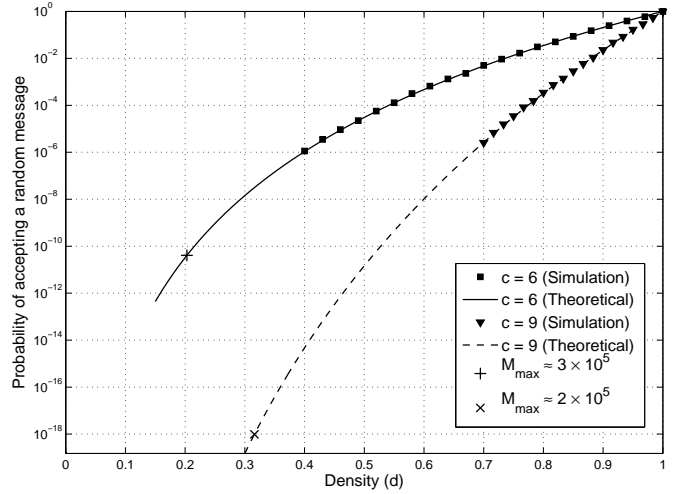

Fig. 5. Type II error rate as a function of the density of the network composed of $\chi = 100$ clusters of $l = 64$ fanals each, with cliques of order $c = 6$ or 9. Densities corresponding to efficiency-1 diversities ($M_{\max}$) are also indicated.

## VI. BLURRED MESSAGES

Because the proposed network uses cliques as support of information and since cliques are redundant structures, the error correction of distorted or blurred incident messages is possible. In order to illustrate this property, we consider a network with parameters $\chi = 100$, $l = 64$ and $c = 12$. $M$ messages are learnt using $c$ contiguous clusters, the first one being randomly located at a position between 1 and $\chi - c + 1 = 89$. After the learning phase, messages are presented to the network after systematic permutation of two consecutive characters, in a cyclic way. For instance, if the messages were, among other things, 12-letter English words, the learned word "intelligence" may be received either as "nietllgineec" or "etnleilegcni". If the network is able to recognize "intelligence" from these pairwise permuted versions, it would also be able to correct less disturbed words, such as "intleligecne" or "intellgienec" (a task that the human brain is also reputed to be easily capable of).

In order to allow the network to cope with the upset order of characters, the initialization of the retrieving process has to be modified: when a fanal is activated in a cluster, then the equivalent fanals (i.e. fanals with the same index values $j$) of the adjacent clusters have also to be activated. So, at the beginning of the retrieving process, each cluster contains three active fanals. The learned clique "intelligence" being contained in the activated sub-graph composed of the $12 \times 3 = 36$ nodes, the decoding procedure described by (12) and (13) will hopefully, after several iterations, switch off the irrelevant fanals and display the word with the right order.

The probability for the decoder to fail is again related to the existence of spurious cliques. The most likely are those who share $c-1$ vertices with the learned clique and use one wrong fanal in the remaining cluster as the last vertex. For a given set of $c-1$ correct vertices, the probability that such a spurious clique exists is $d^{c-1}$. Because there are $2c$ possible patterns having $c - 1$ correct vertices and a wrong one in the last cluster, the probability that a false clique may exist, giving directly the error probability in the retrieving process, is

$$P_e = 1 - \left(1 - d^{c-1}\right)^{2c} \approx 2cd^{c-1} \text{ for } d \ll 1. \quad (22)$$

Fig. 6 exhibits the error rate obtained by simulation after one and six iterations, as well as the theoretical curve deducted from (22). Because messages that are presented to the network are strongly distorted, several iterations are required to approach the theoretical performance. By comparing abscissas in Fig. 3 and Fig. 6, we note that the network is able to correct wrong messages as easily as, even a little better than, erasures. If full anagrams, instead of contiguous permutations, were considered, $c$ fanals would have to be activated in each cluster at the initialization step, instead of only 3. The error probability in deciphering these anagrams would then be given by

$$P_e = 1 - \left(1 - d^{c-1}\right)^{(c-1)c} \approx (c-1)cd^{c-1} \text{ for } d \ll 1. \quad (23)$$

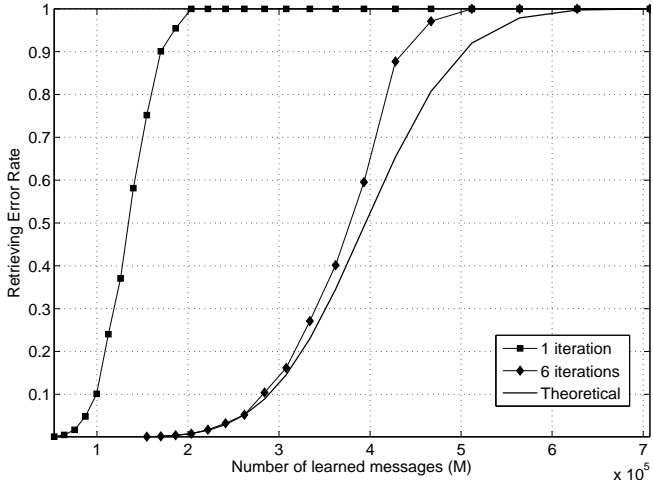

Fig. 6. Error rate in the retrieving of messages distorted by permutations of contiguous characters. The network parameters are $\chi = 100$ and $l = 64$, and messages have order $c = 12$.

## VII. VARIABLE ORDER MESSAGES

We want now to assess the ability of the network with parameters $\chi$ and $l$ to learn and retrieve messages with various values of $c$. From Fig. 4, we can notice that a relatively large set of orders $c$ around $c_{opt}$ may be employed without much deteriorating the retrieving performance of the network. For instance, with a targeted error rate $P_0$ equal to $10^{-4}$ and for this particular network with $\chi = 100$ and $l = 64$, values of $c$ chosen between 8 and 20 would decrease the diversity by about 10% or less from the optimal value obtained with $c_{opt} = 12$.

To formalize the general case, let us consider clique orders distributed between $c_{min}$ and $c_{max}$, such that $1 < c_{min} \leq c_{max} < \chi$. Following the same rationale that led to relation (5), we can express the density of the network as

$$d = 1 - \prod_{c=c_{min}}^{c_{max}} \left(1 - \frac{c(c-1)}{\chi(\chi-1)l^2}\right)^{M_c} \quad (24)$$

where $M_c$ is the number of messages learned with order $c$. If $c$ is uniformly distributed between $c_{min}$ and $c_{max}$, $M_c$ is equal to $\frac{M}{\lambda}$, $M$ still being the total number of i.i.d. messages and $\lambda = c_{max} - c_{min} + 1$.

The learning and retrieving rules are still those given in section III, with threshold values $\sigma_i$, all equal or not, depending on the application. The amount of binary information borne by a particular message $m \in \mathcal{M}$, materialized by a clique with order $c(m)$, is now

$$\log_2\left(\binom{\chi}{c(m)}\right) + c(m)\kappa + \log_2(\lambda).$$

Compared to the writing of (9), a third term has been added to take the choice of $c$ into account, though of minor importance for typical values of $\chi$, $l$, $c_{min}$ and $c_{max}$. Efficiency $\eta$ is then given by

$$\eta = \frac{2\sum_{m \in \mathcal{M}}\left(\log_2\left(\binom{\chi}{c(m)}\right) + c(m)\kappa + \log_2(\lambda)\right)}{\chi(\chi-1)l^2}. \quad (25)$$

It is not easy to exploit this formula in the general case. For a uniform distribution of orders $c$ between $c_{min}$ and $c_{max}$, this amounts to a more convenient formula:

$$\eta = \frac{2M \sum_{c=c_{min}}^{c_{max}}\left(\log_2\left(\binom{\chi}{c}\right) + c\kappa + \log_2(\lambda)\right)}{\lambda\chi(\chi-1)l^2}. \quad (26)$$

As in section III.A, it is then possible to find the efficiency-1 diversity by setting $\eta = 1$. This gives

$$M_{max} = \frac{\lambda\chi(\chi-1)l^2}{2\sum_{c=c_{min}}^{c_{max}}\left(\log_2\left(\binom{\chi}{c}\right) + c\kappa + \log_2(\lambda)\right)}. \quad (27)$$

For instance, taking $c_{min} = 12$ and $c_{max} = 20$ gives $M_{max} = 1.27 \times 10^5$, a value very close to that obtained for $c$ constant and equal to 16, as seen in section III.a.

Finally, to finish this survey of sparse messages stored in network of neural cliques, we consider a network with parameters $\chi$ and $l$ learning $M$ i.i.d. messages of variable and uniformly distributed orders $c$ and the blind recovery of which is done without the knowledge of $c_e = \alpha c$ sub-messages. Relation (15) being applicable to each of the subset of the messages with particular order $c$, we can write the average retrieving error rate as

$$\overline{P_e} \approx \frac{1}{\lambda} \sum_{c=c_{min}}^{c_{max}} \left(1 - \left(1 - d^{(1-\alpha)c}\right)^{\alpha c(l-1)+(\chi-c)l}\right). \quad (28)$$

Combined with (24) for $M_c = \frac{M}{\lambda}$, it is then possible to estimate the performance of the network in terms of error rate versus diversity when variable order messages are learned. Fig. 7 shows this performance in two cases: $c_{min} = 6$ and $c_{max} = 18$ on the one hand, $c_{min} = 12$ and $c_{max} = 24$ on the other hand, and for $\alpha = 0.25$ (on average when $c$ is not a multiple of 4). Both simulated and theoretical values deducted from (28) are also displayed. These curves have to be compared with that obtained for $c$ constant and equal to 12 (Fig. 3(a)). Unsurprisingly, the performance is somewhat lower for variable order messages than for $c$ constant and close to $c_{opt}$. For low error rates, reduction in diversity is more pronounced for small values of $c$. But in all cases, provided that $c$ remains small compared to $\chi$, the order of magnitude for diversity remains the same.

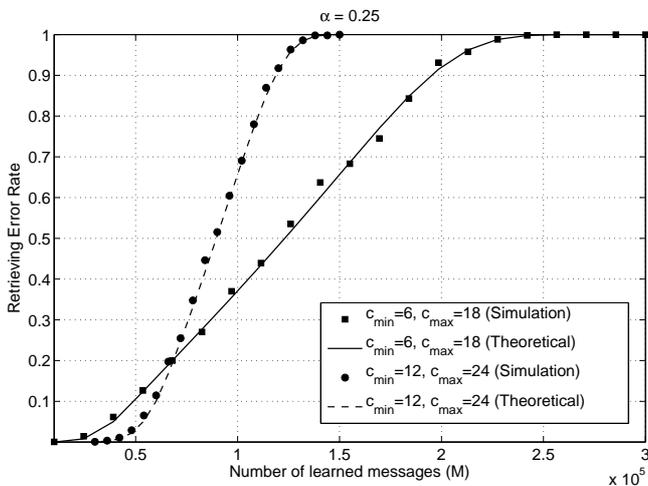

Fig. 7. Blind recovery error rate of messages with variable order between $c_{min}$ and $c_{max}$ in a network with parameters $\chi = 100$ and $l = 64$, and when $c_e = c/4$ clusters (on average) have no initial information.

## VIII. CONCLUSION

We have demonstrated and assessed the ability of binary networks to learn and retrieve a large number of sparse messages, of constant or variable orders. These messages are stored as cliques whose vertices belong to distinct clusters, the number of which ($c$) is small compared to the total number ($\chi$). The diversity (the number of learnable messages) is proportional to the squared number of nodes. The learned messages may be retrieved in presence of erasures and even after some kind of distortion, provided that the decoding algorithm is adapted to the particular problematic.

To speak in terms of non linear systems, we have shown that an appropriately organized binary recurrent graph may contain a large number of attractors. These stable and distinguishable patterns may be considered as codewords of a distributed code whose local codes are constant weight-1 codes [7] and the global code is a clique-based code. The big difference and advantage of this "neural code" over a classical distributed code, such as turbo [14] or LDPC [15] code, is of course its ability to learn independent codewords, that is, words that are not linked by a linear coding relation.

The construction of the proposed network was inspired by the hierarchical organization of the neocortex: microcolumns (fanals) grouped in columns (clusters) which gather in macrocolumns (networks). We have seen that the number of messages that a network composed of $\chi = 100$ clusters having $l = 64$ fanals each is able to learn and retrieve correctly is around 100,000. If we extrapolate this result to what could offer the resource of the human cortex with its billion microcolumns (roughly), the quadratic law expressed by (8) leads to $10^5 \left( \frac{10^9}{6400} \right)^2 \approx 10^{15}$ messages. This order of magnitude is certainly exaggerated as the "small world" organization of the neocortex [16] does not allow extending the quadratic law to the whole scale.

The concept of "neural clique" is familiar to neuroscientists [17],[18], but to our knowledge, the learning and retrieving properties of clique-based sparse messages had never been studied to the point of formalization we have developed in this paper. In a recent communication [19], such cluster-based networks were also demonstrated to be suited to the learning of sequences, and not only to atemporal messages. To allow this, cliques are replaced with tournaments, that is, cliques where arrows substitute for edges. Efficiencies in the range of 20% are achievable with still good properties of robustness, tournaments being structures almost as redundant as cliques. The kind of neural networks that we have analyzed in this paper has very simple learning and retrieving rules and offers a large amount of storage capacity as well as attractive correction properties. This may be considered as a good candidate for modeling the cerebral long term memory and also an interesting starting point for the design of machines able to learn a lot of messages/situations/sequences and to combine them thanks to some cognitive principles yet to be defined. This work is currently undertaken within the framework of the NEUCOD ("Neural coding") project founded by the European Research Council.

## APPENDIX

Fig. 8 depicts a learned clique of order 6 with vertices A, B, C, D, E and F, all belonging to distinct clusters. Only vertices A, B, C and D are known at the beginning of the retrieving process. A spurious clique of order 5 with vertices A, B, C, D and X exists, due to connections established by other learned cliques (not represented).

So, at the beginning, A, B, C and D have initial values 1 whereas the others have value 0. Then unitary signals are sent through the network and equation (11) with memory parameter $\gamma = 1$ fixes all values to 4. Spurious vertex X cannot be eliminated using relation (12); all values are then positioned to 1. If a second iteration is carried out, the scores become A = B = C = D = 7, E = F = 6 and X = 5. Again, because E and F do not obtain the maximal score which is 7, only nodes A, B, C and D are assigned value 1 and the clique cannot be recovered. Further iterations will repeat the same cycle.

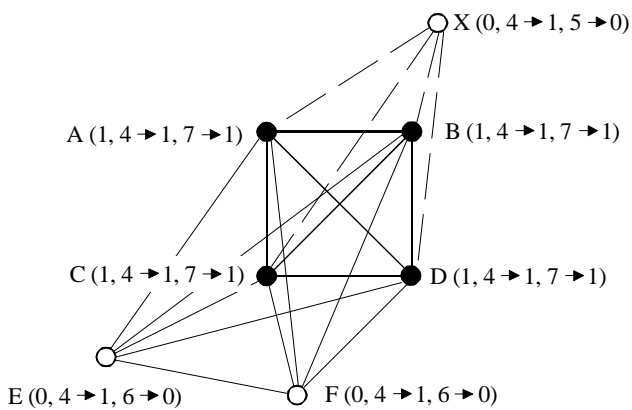

Fig. 8. An example of failure in the retrieving of a learned clique (A-B-C-D-E-F) due to a spurious clique of shorter order (A-B-C-D-X).

**Behrooz Kamary Aliabadi** (S'09) is currently a Ph.D. student at Télécom Bretagne (Institut Mines-Télécom, France). He received his M.Sc. in Wireless Communications from Lund University, Lund, Sweden and his B.Sc. in Telecommunications from Azad University, Tehran, Iran. His research interests include neural networks, information theory and signal processing.

**Claude Berrou** (M'86–F'09) is a Professor in the Electronics Department of Télécom Bretagne (Institut Mines-Télécom, France). In the early 1990s, in collaboration with Prof. A. Glavieux, he introduced the concept of probabilistic feedback into error-correcting decoders and developed a new family of quasioptimal error correcting codes, which he nicknamed "turbo codes." He also pioneered the extension of the turbo principle to joint detection and decoding processing, known today as "turbo detection" and "turbo equalization." His current research interests include computational intelligence in the light of information theory. He has received several distinctions, amongst which are the IEEE (Information Theory) Golden Jubilee Award for Technological Innovation in 1998, the IEEE Richard W. Hamming medal in 2003, the Grand Prix France Télécom de l'Académie des Sciences in 2003, and the Marconi Prize in 2005. He was elected as a member of the French Academy of Sciences in 2007.

**Vincent Gripon** (S'10-M'12) obtained his Ph.D. degree at Télécom Bretagne (Institut Mines-Télécom, France). He is currently a post-doc at McGill University, Montréal, Canada. He is the co-creator and organizer of a programming contest named TaupIC, which targets French top undergraduate students. His current research interests include information theory, error-correcting codes and cognitive science, and links between neural networks and distributed error-correcting codes.

**Xiaoran Jiang** (S'11) was born in Hangzhou, China, in 1987. He obtained his engineer degree (Master equivalent) in Telecommunication from Télécom Bretagne (Institut Mines-Télécom, France) in 2010. He is currently a Ph.D student with the Electronics department of Télécom Bretagne. His current research interests include information theory, sparse coding, cognitive science, and especially the sequence learning in neural networks.